\documentclass[letterpaper, conference]{ieeeconf}
\IEEEoverridecommandlockouts
\usepackage{cite}

\def\BibTeX{{\rm B\kern-.05em{\sc i\kern-.025em b}\kern-.08em
    T\kern-.1667em\lower.7ex\hbox{E}\kern-.125emX}}
    
  \usepackage{amsmath,amssymb,amsfonts,amsthm} 
\usepackage{mathtools}
\usepackage{commath}
\usepackage{algorithmic}
\usepackage[ruled,vlined]{algorithm2e}
\usepackage{graphicx}
\usepackage{textcomp}
\usepackage{xcolor}

\usepackage{framed}
\definecolor{shadecolor}{rgb}{.9,.9,.9}
\usepackage{tcolorbox}
\usepackage{siunitx}
\usepackage{textgreek}
\usepackage{url}
\usepackage{booktabs, tabularx}
\usepackage{multirow}
\usepackage{stackengine}
\usepackage[caption=false, font=footnotesize]{subfig}

\usepackage{tikz}
\usetikzlibrary{quotes,angles}
\usetikzlibrary{calc}
\usetikzlibrary{decorations.pathmorphing}

\usepackage{soul}

\usepackage[absolute]{textpos}     
\usepackage[hidelinks]{hyperref}   

\title{\LARGE\bf Docking and Undocking a Modular Underactuated Oscillating Swimming Robot
}

\author{Gedaliah Knizhnik and Mark Yim
\thanks{The authors are with the GRASP Laboratory, University of Pensylvannia, Philadelphia, PA 19104. 
        {\tt\footnotesize knizhnik@seas.upenn.edu}}%
}

\newcommand{\copyrightstatement}{
    \begin{textblock*}{5.7in}(0.25in,0.25in) 

        \noindent
        \footnotesize
        This accepted article to ICRA is made available by the authors in compliance with IEEE policy.

        \noindent
        Please find the final, published version in IEEE Xplore, DOI: \href{https://doi.org/10.1109/ICRA48506.2021.9562033}{\textcolor{blue}{10.1109/ICRA48506.2021.9562033}}.

    \end{textblock*}

    \begin{textblock*}{5.7in}[0,1](0.25in,10.85in) 

        \noindent
        \footnotesize \scriptsize
        \copyright 2021 IEEE. Personal use of this material is permitted.
        Permission from IEEE must be obtained for all other uses, in any current or future media, including reprinting/republishing this material for advertising or promotional purposes, creating new collective works, for resale or redistribution to servers or lists, or reuse of any copyrighted component of this work in other works.
    \end{textblock*}
}

\newtheorem{definition}{Definition}

\def\centerarc[#1](#2)(#3:#4:#5)
    { \draw[#1] ($(#2)+({#5*cos(#3)},{#5*sin(#3)})$) arc (#3:#4:#5); }
    
\def\drawboat(#1:#2:#3)(#4)(#5:#6){
    
    \def\radius{1.5}
    \def\radiusTop{1.6*\radius}
    \def\radiusTail{1.75*\radius}
    \def\leftAngle{135}
    \def\rightAngle{45}
    \def\tailLeft{180}
    \def\tailRight{0}
    
    \coordinate (O) at #1;

    \begin{scope}[rotate around={#2:(O)}]
        
        \ifthenelse{\equal{#6}{1}}{
            \fill (O) circle[radius=2pt] node[below left] {\LARGE #5};
        }{
            \fill (O) circle[radius=2pt] node[below left] {#5};
        }
        
        \begin{scope}[rotate around={#3:(O)}]
        
            \path (O) +(\leftAngle:\radius) coordinate (FL);
            \path (O) +(\rightAngle:\radius) coordinate (FR);
            \path (O) --++ (0,-\radiusTail) coordinate (T) {};
            \path (O) +(\tailLeft:\radius) coordinate (TL);
            \path (O) +(\tailRight:\radius) coordinate (TR);
            
            \begin{scope}[rotate around={90:(FL)}]
            \draw[ultra thick, gray] (FL) arc (0:50:1.5) coordinate (FLEnd);
            \end{scope}
            
            \begin{scope}[rotate around={90:(FR)}]
            \draw[ultra thick, gray] (FR) arc (0:-50:1.5) coordinate (FREnd);
            \end{scope} 
            
            \centerarc[blue, ultra thick](O)(\tailLeft:\tailRight:\radius);
            \draw[blue, ultra thick] (TL) to[bend right] (T);
            \draw[blue, ultra thick] (TR) to[bend left] (T);
            
        \end{scope}
        
        \draw[very thick] (O) circle[radius=\radiusTop];
        \path (O) +(90:\radiusTop) coordinate (Front);
        \draw[dashed, thick, ->] (O) -- (Front);
        
        \path (O) +(0:\radiusTop) coordinate (MR);
        \path (O) +(90:\radiusTop) coordinate (MT);
        \path (O) +(180:\radiusTop) coordinate (ML);
        \path (O) +(270:\radiusTop) coordinate (MB);
        
        \ifthenelse{\equal{#4}{1}}{        
                \fill (MR) circle[radius=2pt] node[left] {\Large $-\frac{\pi}{2}$};
                \fill (MT) circle[radius=2pt] node[above] {\Large $0$};
                \fill (ML) circle[radius=2pt] node[left] {\Large $\frac{\pi}{2}$};
                \fill (MB) circle[radius=2pt] node[above] {\Large $\pi$};
                }{}
        \ifthenelse{\equal{#4}{2}}{        
                \fill (MR) circle[radius=2pt] node[right] {};
                \fill (MT) circle[radius=2pt] node[above] {};
                \fill (ML) circle[radius=2pt] node[left] {};
                \fill (MB) circle[radius=2pt] node[above right] {};
                }{}
    
    \end{scope}
}

\begin{document}
\bstctlcite{MyBSTcontrol} 
\copyrightstatement                    

\maketitle

\begin{abstract}
We describe a docking mechanism and strategy to allow modular self-assembly for the Modboat: an inexpensive, underactuated, oscillating, surface-swimming robot powered by a single motor. Because propulsion is achieved through oscillation, orientation can be controlled only in the average; this complicates docking, which requires precise position and orientation control. Given these challenges, we present a docking strategy and a motion primitive for controlling orientation, and show that this strategy allows successful docking in multiple configurations. Moreover, we demonstrate that the Modboat is also capable of undocking and changing its dock configuration, all without any additional actuation. This is unique among similar modular robotic systems.
\end{abstract}


\section{Introduction} \label{sec:intro}

Modular and self-reconfigurable aquatic systems show great potential for versatility and utility in modern applications. Individual modules can be made inexpensive, and various capabilities can be introduced through cooperation and the modular structure, rather than through the units. This can enable significantly less expensive and more versatile robots. A single aquatic robot may carry a single flow sensor, for example, providing little utility. But a system of robotic units attached together, each carrying a flow sensor, can provide a detailed picture of the flow field in a particular region with a precise spatial distribution. 

Very little work, however, has been done on modular and self-reconfigurable robots in aquatic applications. The TEMP project explored building structures from rectangular robotic modules that could dock together~\cite{OHara2014}\cite{Paulos2015}, and the Roboats project considered docking multiple surface modules ~\cite{Wang2018DesignVehicle}\cite{Marques2019DecentralizedEnvironments}. We are not aware of other docking systems for aquatic surface operations, however, and both of these systems involve expensive, holonomic units. Underneath the water's surface, AMOUR has demonstrated docking capabilities~\cite{Vasilescu2010AMOURPayloads}, as has ANGELS~\cite{Mintchev2014TowardsRobots}, but both of these systems can reconfigure in only 1D. CoCoRo~\cite{Mintchev2014} has also demonstrated docking capabilities~\cite{Mintchev2014TowardsRobots}, but these have been used exclusively for connecting to a fixed power station, not other modules. 

\begin{figure}[t]
    \centering
    \resizebox{0.85\linewidth}{!}{
        \begin{tikzpicture}[pencildraw/.style={
        black!75,
        decorate,
        decoration={random steps,segment length=0.8pt,amplitude=0.5pt}
        }]
        
        \coordinate(O) at (0,0);

        \drawboat((O):0:0)(1)(T:0);
        \def\aoaAngle{45};
        \def\aoaRad{2*\radiusTop}
        \def\arcRad{\radiusTop + \aoaRad}
        
        \path (O) +(\aoaAngle:\radiusTop) coordinate(AOL);
        \path (O) +(-\aoaAngle:\radiusTop) coordinate(AOR);
        
        \path (AOL) +(\aoaAngle:1.6) coordinate(AOAL);
        \path (AOR) + (-\aoaAngle:1.6) coordinate(AOAR);
        
        \draw[pencildraw, dashed, ultra thick, green] (AOL) to (AOAL);
        \draw[pencildraw, dashed, ultra thick, green] (AOR) to (AOAR);
        \centerarc[pencildraw, dashed, ultra thick, green](O)(-\aoaAngle:\aoaAngle:4);
        
        \path(AOL) +(135:0.2) coordinate(T1);
        \path(AOAL) +(135:0.2) coordinate(T2);
        \draw[<->] (T1) to (T2);

        \coordinate (T3) at ($ (T1) !.5! (T2)$);
        \fill (T3) circle[radius=0pt] node[above left] {$d_{cap}$};
        
        \draw
            pic["$\phi_{cap}$",text=black, draw=black, <->, angle eccentricity=1.15, angle radius=3cm]{angle=AOAR--O--AOAL};

    \end{tikzpicture}}
    \caption{A simplified diagram of the Modboat, with the docking points indicated and labeled and the approximate area of acceptance of the $-\pi/2$ dock (not to scale) in green. The top body is shown in black, the bottom body in blue, and the flippers in gray. We have approximately measured $\phi_{cap} \approx 90^\circ$ and $d_{cap} \approx 15\si{cm}$. This is approximately one diameter of the Modboat top body. The other dock points have similar areas of acceptance by symmetry.}
    \label{fig:areaAccept}
\end{figure}
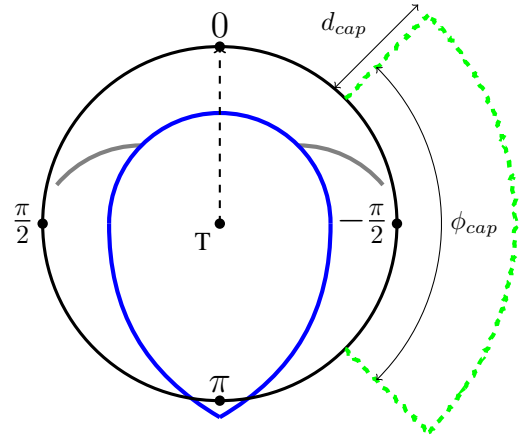

In prior work~\cite{Knizhnik2020} we introduced the Modboat: an inexpensive, oscillating, surface-swimming robot powered by a single actuator and ready for modular self-assembly; passive flippers convert the rotation of its two concentric bodies (see Fig.~{\ref{fig:areaAccept}}) into a forward translation. We showed that pauses in the oscillation could be used to steer a Modboat towards a target as a discrete-time single-integrator~\cite{Knizhnik2020PausesRobot}. Critically, all parts of the Modboat oscillate in order to achieve propulsion, and while swimming its orientation can be controlled only \textit{in the average}. This makes docking --- a process involving precise control of both position and orientation --- very difficult. It is not sufficient to plan a path accounting for the propulsion-orientation coupling as with other non-holonomic systems.

In this work, we present a mechanism and strategy that allows a Modboat to dock to/undock from other units without additional actuation. This is unique, as all other similar systems require additional actuators for docking and latching~\cite{OHara2014, Paulos2015,Wang2018DesignVehicle,Marques2019DecentralizedEnvironments, Vasilescu2010AMOURPayloads,Mintchev2014TowardsRobots,Mintchev2014}. We also describe a novel motion-primitive for controlling the instantaneous orientation, which allows a choice of docking configurations. This allows the Modboat to self-assemble and reconfigure in 2D with an affordable and easily scalable system. 

The rest of this work is organized as follows. We discuss the docking mechanism and the strategy for using it in Sections~\ref{sec:dockingMech} and~\ref{sec:docking}, respectively. In Section~\ref{sec:oriControl} we discuss a new motion primitive for controlling instantaneous orientation, and in Section~\ref{sec:undocking} we demonstrate how the tail can be used for undocking. We experimentally demonstrate docking in Section~\ref{sec:experiments} and discuss the results in Section~\ref{sec:discussion}.


\section{Docking Mechanism} \label{sec:dockingMech}

The Modboat is modeled as moving in the plane and is composed of two concentric bodies, as seen in Fig.~\ref{fig:areaAccept}. The bottom body provides a mounting point for the flippers, which rotate freely, and for the tail. The top body is larger and contains the control hardware; it is the \textit{primary} body of the Modboat. The robot weighs $0.66\si{kg}$, and the top body has a diameter of $15.2\si{cm}$. A full description of the design can be found in our prior work~\cite{Knizhnik2020}. Note that the flippers do not protrude from the top body footprint, which prevents them from mechanically interfering with neighboring docked boats. The tip of the tail, however, does protrude. The purpose for this will be discussed in Section~\ref{sec:undocking}.

Permanent magnets are advantageous for low-cost docking applications as they can passively align and guide the modules when docking. The resulting dock consumes no energy when active, which is also advantageous due to limited onboard energy capacity. Magnetic docks also provide a reasonably large area of acceptance. It should be noted, however, that such passive docks will also present control challenges to modules attempting to operate in close proximity without docking, as they cannot be switched off.

Four permanent magnets are therefore placed at the four cardinal points inside the top body, as shown in Fig.~\ref{fig:areaAccept}. This allows the circular Modboat to form a square lattice when docked. Each Modboat is assigned either a magnetic $N$ or $S$ designation, and each magnet is placed so that the designated pole faces outward; this gives the boats polarity rather than the docks. A valid square lattice can easily be formed from such polarized boats (consider a chess-board), and it is reasonable in large self-assembly applications to assume an infinite number of modules of either polarity. Nongendered/hermaphroditic magnetic docking setups are possible, but they show lower areas of acceptance\cite{Eckenstein2019High-DimensionalGeometries}.

Each magnet is rated for a pull strength of $67\si{N}$; two layers of $1/8\si{in}$ acrylic enclosing each boat and the air gaps due to flat magnets in cylindrical shells result in an effective docking strength of $4.1\si{N}$. Each Modboat weighs $0.66\si{kg}$ (or $6.5\si{N}$), and this docking strength has been experimentally verified to be sufficient to hold two boats together when swimming. 

The area of acceptance of a magnetic dock depends greatly on the fluid and robot velocities, as well as on magnetic interactions between the four magnets on each boat involved. This makes it difficult to explicitly quantify. We roughly approximate the area of acceptance, however, as a $90^\circ$ wide cone extending to $\approx 15\si{cm}$ from the boat's edge, which is shown in Fig.~\ref{fig:areaAccept}; another magnet entering this region will be captured. This assumes the fluid and the boats are approximately at rest, and the boats are approximately aligned in orientation\footnote{In fact, our experiments show that even a small velocity pointed away drastically reduces the area of acceptance.}. This distance can be considered the minimum safe distance for operating individual Modboats in proximity to one another. 

We define a \textbf{dock} by its orientation $\psi \in \{-\pi/2, 0, \pi/2, \pi \}$ relative to the "front" of the Modboat (at $0$) as shown in Fig.~\ref{fig:areaAccept}, and by a subscript identifying each boat. The orientation of the Modboat top body in the world-frame\footnote{In our previous work~\cite{Knizhnik2020}\cite{Knizhnik2020PausesRobot} we use $\theta$ to refer to the orientation of the bottom body and $\theta_t$ to refer to the top body orientation. We use $\theta$ here for the top body to avoid unnecessary subscripts, as only the top body orientation is relevant in this work.} is given by $\theta$.

Since the dock directions are given relevance by the Modboat's preferred front direction, we want to be able to choose which pair of docks are used on each side. The method for doing so is discussed in Sections~\ref{sec:dockingStrat} and~\ref{sec:oriControl}, and we define:

\begin{definition} \label{def:front}
A \textbf{front-dock} involves the $\psi_{b} = 0$ dock.
\end{definition}
\begin{definition} \label{def:side}
A \textbf{side-dock} involves the $\psi = \pm \pi/2$ docks.
\end{definition}
\begin{definition} \label{def:rear}
A \textbf{rear-dock} involves the $\psi = \pi$ dock.
\end{definition}


\section{Docking Strategy} \label{sec:docking}

Conventional techniques for self-assembly generally add single units to the assembled whole rather than combining large sub-assemblies~\cite{Seo2013AssemblyRobots}.  This is often sensible from a dynamics perspective, since individual modules are often far more mobile than a conglomerate. Moreover, the dynamics of multiple Modboats swimming together have not been explored to date, so we consider docking as defined in Definition~\ref{def:docking}, with nomenclature as in Definitions~\ref{def:mobile} and~\ref{def:target}.

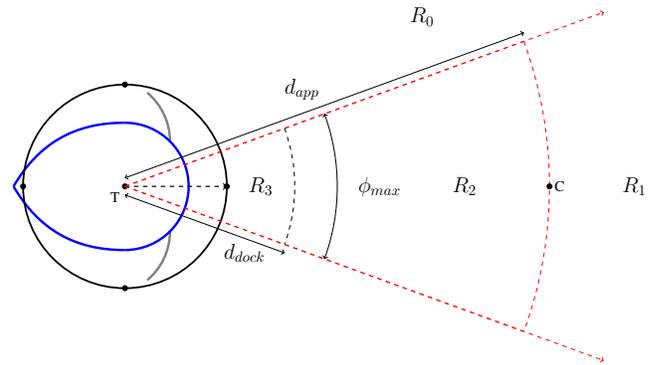
\begin{figure}[t]
    \centering
    \resizebox{\linewidth}{!}{ \begin{tikzpicture}
        \def\distToApp{10}
        \def\distToDock{4}
        \def\angTol{20}
        
        \coordinate(O) at (0,0);
        
        \drawboat((O):-90:0)(2)(T:0);
        
        \path (O) +(\angTol:\distToApp) coordinate (A);
        \path (O) +(-\angTol:\distToApp) coordinate (B);
        
        \draw[red, dashed] (O) to (A);
        \draw[red, dashed] (O) to (B);
        
        \centerarc[red, dashed](O)(\angTol:-\angTol:\distToApp);
        
        \path (O) +(0:\distToApp) coordinate (C);
        \fill (C) circle[radius=2pt] node[right] {C};
        
        \centerarc[black, dashed](O)(\angTol:-\angTol:\distToDock);
        
        \path (A) +(\angTol:0.2*\distToApp) coordinate (A1);
        \path (B) +(-\angTol:0.2*\distToApp) coordinate (B1);
        
        \draw[red, dashed, ->] (O) to (A1);
        \draw[red, dashed, ->] (O) to (B1);
        
        \path (O) +(0:1.2*\distToApp) coordinate (E);
        \fill (E) circle[radius=0pt] node {\Large $R_1$};
        
        \path (O) +(0:8) coordinate (F);
        \fill (F) circle[radius=0pt] node {\Large $R_2$};
        
        \path (O) +(0:3.2) coordinate (G);
        \fill (G) circle[radius=0pt] node {\Large $R_3$};
        
        \coordinate (H) at (7,4);
        \fill (H) circle[radius=0pt] node {\Large $R_0$};
        
        
        \path(O) + (90:0.2) coordinate(O2);
        \path(A) + (90:0.2) coordinate(A2);
        \draw[black, <->] (O2) to (A2);
        
        \coordinate (A2L) at ($ (O2) !.5! (A2)$);
        \fill (A2L) circle[radius=0pt] node[above left] {\Large $d_{app}$};
        
        \path(O) + (90:-0.2) coordinate(O3);
        \path(O) + (-\angTol:\distToDock) coordinate(B2);
        \path(B2) + (90:-0.2) coordinate(B3);
        \draw[black, <->] (O3) to (B3);
        
        \coordinate (B3L) at ($ (O3) !.75! (B3)$);
        \fill (B3L) circle[radius=0pt] node[below] {\Large $d_{dock}$};
        
        \draw
            pic["\Large $\phi_{max}$",text=black, draw=black, <->, angle eccentricity=1.2, angle radius=5cm]{angle=B--O--A};
    
    \end{tikzpicture}    }
    \caption{A diagram of the docking strategy regions used by a mobile boat (not shown) to dock to the target boat $T$. The strategy is shown in detail in Fig.~\ref{fig:dockStratFSM}. Region dimensions are not to scale.}
    \label{fig:dockStrat}
\end{figure}

\begin{definition} \label{def:docking}
The \textbf{docking problem} is guiding a single swimmer from initial location $(x_0,y_0)$, orientation $\theta_0$, and desired dock $\psi_b$ to dock with a target at location $(x_t,y_t)$, orientation $\theta_t$, and desired dock $\psi_t$. The target cannot swim or adjust its orientation.
\end{definition}

\begin{definition} \label{def:mobile}
The \textbf{mobile boat} is the active swimmer being guided in the docking problem.
\end{definition}

\begin{definition} \label{def:target}
The \textbf{target boat} is the boat to which the mobile boat is docking. The target boat may not be fixed but is assumed to be stationary.
\end{definition}

A single targeted Modboat may be able to adjust its orientation to assist in the dock, simplifying the trajectory for the mobile boat. A target that is docked to other units, however, cannot do so. Thus the mobile boat is tasked with achieving both the position and orientation necessary for docking.


\subsection{Strategy} \label{sec:dockingStrat}

When swimming, the Modboat is controlled by a heading from it to a target waypoint. We create a docking strategy by using a discrete and non-holonomic approximation of the attractive well used by Saldana et. al\cite{Saldana2018ModQuad:Midair}. This strategy can be summarized in three stages:

\begin{enumerate}
    \item \textbf{Distancing:} Achieve distance from the target so that a perpendicular approach towards the desired dock is available.
    \item \textbf{Homing:} Approach the target along the perpendicular to establish a drift velocity toward the desired dock. 
    \item \textbf{Orienting:} Control the orientation to present the desired dock on the mobile boat to the desired dock on the target boat.
\end{enumerate}

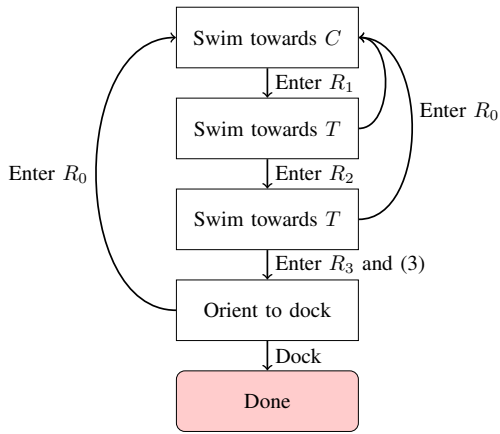
\begin{figure}[t]
    \centering
    \resizebox{0.8\linewidth}{!}{
        \begin{tikzpicture}[node distance=1.5cm]
    
        \tikzstyle{startstop} = [rectangle, rounded corners, minimum width=3cm, minimum height=1cm,text centered, draw=black, fill=red!20]
        
        \tikzstyle{process} = [rectangle, minimum width=3cm, minimum height=1cm, text centered, draw=black]
        \tikzstyle{arrow} = [thick,->]

        \node (pro1) [process] {Swim towards $C$};
        \node (pro2) [process, below of=pro1] {Swim towards $T$};
        \node (pro25) [process, below of=pro2] {Swim towards $T$};
        \node (pro3) [process, below of=pro25] {Orient to dock};
        \node (stop) [startstop, below of=pro3] {Done};
        
        \draw [arrow] (pro1) -- node[anchor=west] {Enter $R_1$} (pro2);
        \draw [arrow] (pro2) -- node[anchor=west] {Enter $R_2$} (pro25);
        \draw [arrow] (pro25) -- node[anchor=west] {Enter $R_3$ and~\eqref{eq:tranToOri}} (pro3);
        \draw [arrow] (pro3) -- node[anchor=west] {Dock} (stop);
        
        \draw [arrow] (pro2) edge[bend right=90] node[anchor=west] {} (pro1);
        \draw [arrow] (pro25) edge[bend right=90] node[anchor=south west] {Enter $R_0$} (pro1);
        \draw [arrow] (pro3) edge[bend left=90] node[anchor=east] {Enter $R_0$} (pro1);
        
        \end{tikzpicture}                          
    }
    
    \caption{An illustration of the docking procedure followed by the mobile boat. The procedure moves linearly unless an abort is triggered, at which point it resets.}
    \label{fig:dockStratFSM}
\end{figure}

Fig.~\ref{fig:dockStrat} demonstrates the principal elements of this docking strategy, which are implemented as a finite state machine as shown in Fig.~\ref{fig:dockStratFSM}. From an initial state in $R_0$ the mobile boat is driven towards a virtual waypoint at $C$ defined by a distance $d_{app}$, the orientation of the target $\theta_t$, and the desired target dock $\psi_t$. This serves to establish an (almost) perpendicular approach to the target dock. 

Once it is within $R_1$ (i.e. within the cone and further than $d_{app}$ from the target), the mobile boat is directed towards the target at $T$. This waypoint is maintained within both $R_1$ and $R_2$, serving to establish a drift velocity towards the target. 

Finally, within $R_3$ the mobile boat orients itself to present the desired dock $\psi_b$ using the method described in Section~\ref{sec:oriControl}. The Modboat cannot simultaneously control its orientation and the direction of its translation, so it relies on the previously established velocity to drift the rest of the way to the target while the orientation is being set. The transition to $R_3$ is based both on a distance criterion (being closer to the target than $d_{dock}$) and an angular velocity criterion that will be discussed in Section~\ref{sec:oriControl}.

Because the Modboat is underactuated, we insist that transitions occur \textit{in order} without skipping. Skipping $R_1$, for example, could create a scenario where the drift velocity was insufficiently directed at the target for a successful dock. However, if at any point the Modboat returns to $R_0$ the dock attempt is aborted and the procedure resets.

Note that this docking procedure assumes that motion occurs only in the plane. We have not accounted for disturbances normal to the plane, such as waves. Magnetic docks provide some compensation for height misalignment, but full consideration of out-of-plane motion is left for future work.


\section{Orientation Control} \label{sec:oriControl} 

In prior work~\cite{Knizhnik2020PausesRobot} we focused on controlling the \textbf{heading} of the Modboat, defined at discrete time intervals as the average orientation over a \textbf{cycle}. Self-assembly, however, requires that the \textit{instantaneous} orientation of the boat matches that of the target. We therefore need to bridge the transition from swimming --- treated discretely --- to docking, which must be treated continuously.

An additional motion primitive to control the instantaneous orientation provides this bridge. We can use the bottom body as a reaction wheel to control the orientation of the top body. Because angular acceleration activates the flippers and induces translation, this motion primitive is of limited use, but we theorize that it can be useful when:

\begin{enumerate}
    \item The orientation can be controlled \textbf{slowly}, allowing slow motion that does not activate the flippers \label{item:1} OR
    \item The orientation needs to be controlled only \textbf{briefly}, so that induced translation is unlikely to matter. \label{item:2}
\end{enumerate}

Item~\ref{item:1} may be relevant in docking scenarios where the target boat is free to rotate. It can spend a relatively long time achieving its orientation while the mobile boat approaches. For the docking problem as defined in Definition~\ref{def:docking}, however, we rely instead on item~\ref{item:2}; the orientation control maneuver for the mobile boat will occur in a short period before the dock, minimizing the effect of any induced translation.

Orientation control is therefore achieved by defining a PID controller as in~\eqref{eq:error} and~\eqref{eq:oriPID}, where $\theta$ is the orientation of the top body and $\alpha$ is commanded as angular acceleration to the motor. To minimize flipper activation, we limit the allowable angular velocity of the motor during this mode to $4 \si{rad/s}$.
\begin{align}
    e &= \theta_{d} - \theta \label{eq:error}\\
    \alpha &= K_p e + K_d \frac{de}{dt} + K_i \int e dt \label{eq:oriPID}
\end{align}

It is important to note that --- because oscillations induce translations in the Modboat system --- the orientation controller should be overdamped, but this induces a slow rise time. A long drift period could compensate for this, but it would require higher precision in the approach heading, which we cannot guarantee.

\begin{table}[h]
    \centering
    \caption{Orientation control PID coefficients}
    \begin{tabular}{r|c|c|c} \toprule
    Mode & $K_p$ & $K_i$ & $K_d$ \\ \midrule
    Aggressive & 30 & 0 & 30  \\
    Non-Aggressive & 10 & 0 & 30 \\ \bottomrule
    \end{tabular}
    \label{tab:oriPID}
\end{table}

Since magnetic docking can compensate for angular errors, we therefore choose a gain-scheduled controller model. The initial orientation controller uses an aggressive set of PID coefficients to quickly rise most of the way to the desired orientation. Once the orientation is within $0.15 \si{rad}$ of the target value and the angular velocity is below $1 \si{rad/s}$ we shift to an overdamped non-aggressive controller to prevent oscillations, should the drift period last long enough. The PID coefficients are given in Table~\ref{tab:oriPID}. 

\begin{figure}
    \centering
    \includegraphics[trim = 0cm 0.05cm 0cm 0.30cm, clip=true, width=\linewidth]{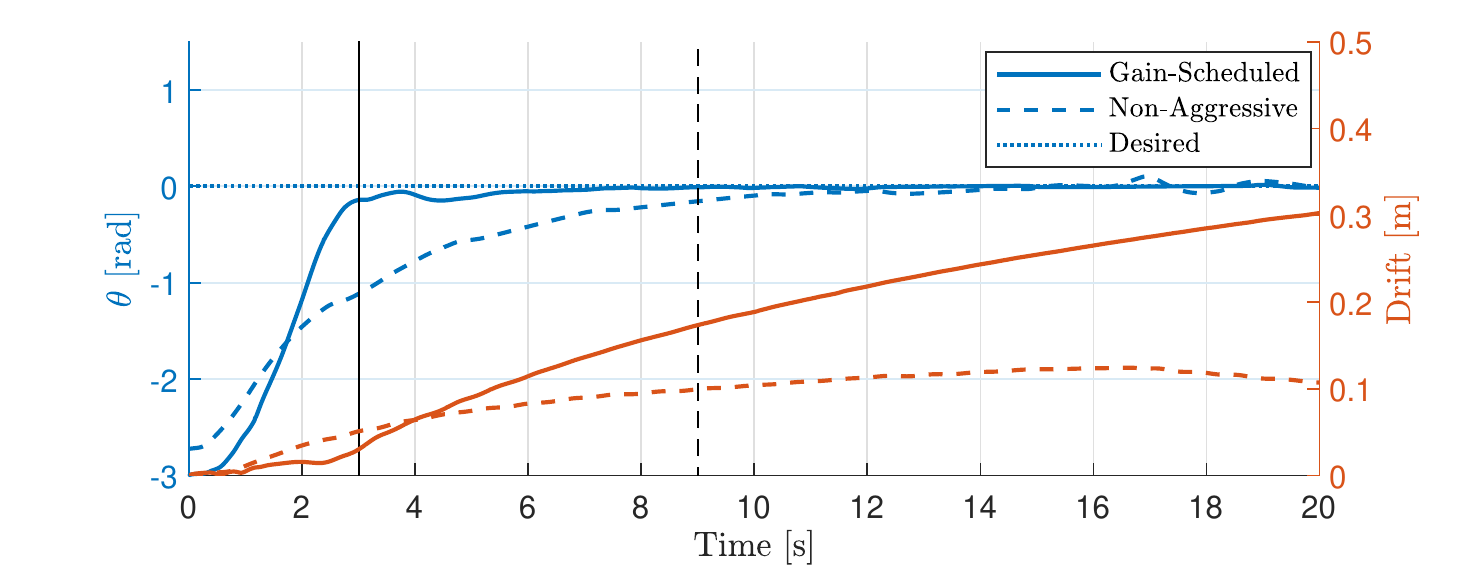}
    \caption{Sample step response of the non-aggressive (blue and orange dashed) and gain-scheduled (blue and orange solid) orientation controllers to a step input of $\approx \pi$ (dotted). Gain-scheduling speeds up rise-time without an appreciable increase in oscillations. This results in more drift overall but is comparable in the initial few seconds, which is sufficient for docking. The vertical lines mark reaching within $0.15\si{rad}$ of the desired orientation.}
    \label{fig:oriControl}
\end{figure}

Figure~\ref{fig:oriControl} shows an example of the performance of the gain-scheduled control approach as compared to using only the non-aggressive controller. The non-aggressive controller takes $9 \si{s}$ to reach the $0.15 \si{rad}$ threshold, while the hybrid controller takes only $3 \si{s}$ without a significant increase in oscillation. We do observe more translation when using the gain-scheduled control approach, but this is offset by the significantly faster correction time when used for docking as per item~\ref{item:2}. Thus, even though the hybrid controller eventually drifts nearly $20\si{cm}$ further, it drifts less in the time needed to rise to the desired orientation.

In order to facilitate docking using orientation control, we add an angular velocity criterion to the strategy in Section~\ref{sec:dockingStrat}. The distance transition criterion to $R_3$ is supplemented with~\eqref{eq:tranToOri}, where $\theta_b$ is the orientation of the mobile boat, $\theta_{des}$ is the desired orientation that will align the mobile and target docks, and the error term is wrapped to $(-\pi,\pi]$. 

This guarantees that the transition occurs when the propulsive oscillations are already turning the Modboat in the desired direction, easing the presentation of the desired dock $\psi_b$. The minimum angular velocity $\omega_{trans}$ ensures this will happen quickly, and the orientation controller then brings this rotation to a halt at the desired angle.
\begin{equation} \label{eq:tranToOri}
    |\dot{\theta}_b| \geq \omega_{trans} \quad \text{and} \quad \dot{\theta}_b (\theta_b - \theta_{des}) < 0 
\end{equation}


\section{Undocking} \label{sec:undocking}

Although we have presented a strategy for docking, modular self-assembly and reconfiguration also requires the ability for modules to undock from each other. In the case of the Modboat, this is achieved without additional actuation by using the tail.

\begin{figure}[t]
    \centering
    \subfloat[\label{fig:docked}]{
        \resizebox{0.45\linewidth}{!}{ \begin{tikzpicture}
            \coordinate(O) at (0,0);
            \drawboat((O):0:0)(2)(A:1);
            \drawboat((4.8,0):0:0)(2)(B:1);
        \end{tikzpicture}}}
          \hfill
    \subfloat[\label{fig:undocked}]{
        \resizebox{0.48\linewidth}{!}{ \begin{tikzpicture}
            \coordinate(O) at (0,0);
            \drawboat((O):0:90)(2)(A:1);
            \drawboat((5.25,0):0:-90)(2)(B:1);
        \end{tikzpicture}}}
    \caption{(a) Two Modboats $A$ and $B$ docked together, with $\psi_A = -\pi/2$ and $\psi_B = \pi/2$. (b) To undock, $A$ would rotate its tail by $\pi/2\si{rad}$ and $B$ would rotate its by $-\pi/2\si{rad}$, forcing the boats apart.}
    \label{fig:dockedBoats}
\end{figure}
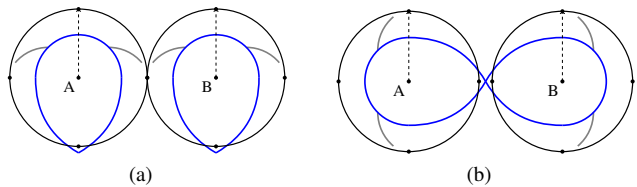

Fig.~\ref{fig:areaAccept} shows a simplified diagram of the Modboat, in which the tail is shown as an extension of the bottom body. The flippers are designed to sit within the footprint of the top body to prevent mechanical interference with neighboring modules~\cite{Knizhnik2020}. The tail, on the other hand, is designed to protrude at its peak, with its curve parametrized so that distance to the center of rotation increased linearly with the angle. This means that two neighboring Modboats (such as shown in Fig.~\ref{fig:dockedBoats}) can rotate their tails into each other to mechanically force separation of the magnetic docks.

The motors used have sufficient torque to separate two docked boats, but the separation provided by two tails is insufficient to completely disable the magnetic attraction. Thus a re-dock occurs immediately after separation if insufficient force is applied by the tail.

Considering Fig.~\ref{fig:dockedBoats}, we must examine two cases. Assuming we focus on boat $A$, either

\begin{enumerate}
    \item Boat $B$ is itself docked to an unspecified number of other modules (not shown in the figure), OR \label{it:undocking1}
    \item Boats $A$ and $B$ are the only boats in the structure. \label{it:undocking2}
\end{enumerate}

In case~\ref{it:undocking1}, the presence of additional modules prevents boat $B$ from rotating and increases its effective inertia during the separation from $A$. We have experimentally determined that an angular velocity of $4\si{rad/s}$ is sufficient to guarantee separation in this case, when both $A$ and $B$ rotate their tails through at least $\pi/2 \si{rad}$. Thus, considering Fig.~\ref{fig:dockedBoats} without loss of generality, $A$ rotates its tail by $\pi/2 \si{rad}$ and $B$ rotates its by $-\pi/2\si{rad}$, each at $4\si{rad/s}$, which separates the boats and imparts enough force to cause them to drift apart. 

In case~\ref{it:undocking2}, however, additional energy is spent rotating and moving boat $B$, which lowers the imparted separation distance and prevents successful undocking. In this case, a sufficient strategy is instead for $A$ to initiate a swim with $\phi_0$ (the centerline of the motor rotation~\cite{Knizhnik2020PausesRobot}) given by the dock connected to $B$ (in Fig.~\ref{fig:dockedBoats} this is $\phi_0 = \pi/2$, so that the tail faces $-\pi/2$), while $B$ rotates its tail to face $A$. This swimming motion creates thrust away from $B$ using the flippers, which adds to the tail-based separation and ensures successful undocking. This method can also be used successfully in case~\ref{it:undocking1}.

It is significant to note that in both cases we observed that, while a tail velocity of $1\si{rad/s}$ resulted in an immediate re-dock in the original configuration, a velocity of $2\si{rad/s}$ consistently resulted in a re-dock in a \textit{new} configuration. In case~\ref{it:undocking1} boat $A$ re-docked rotated by $-\pi/2\si{rad}$ in $15$ of $16$ tests while $B$ maintained its orientation, while in case~\ref{it:undocking2} boat $B$ would also rotate by $\pi/2 \si{rad}$.


\section{Experiments} \label{sec:experiments}

Experiments were performed in a $4.5\si{m} \times 3.0\si{m} \times 1.2 \si{m}$ tank of water equipped with an OptiTrack motion capture system that provided planar position, orientation, and velocities at $120 \si{Hz}$. A MATLAB interface recorded the data, calculated the heading and control parameters, and sent commands to the Modboat for course correction. A set of base parameters $\begin{bmatrix} T_1 & T_2 & A & \phi_0 \end{bmatrix} =\begin{bmatrix} 1.5 & 1.5 & 2 & 0 \end{bmatrix}$ was used for \textbf{all} experiments, with pause control applied over these parameters (see~\cite{Knizhnik2020PausesRobot}). Testing was done in calm water with no external currents applied.

To simulate docking to a larger assembly of boats the target boat was fixed to the bottom of the tank. This target was used for the majority of testing, but we also evaluated docking with a single free-floating non-actuated target. This was intended to evaluate if significant differences exist between the two cases, and if the docking strategy would function for a target that could be pushed away by the water moving with the incoming mobile boat.

\begin{figure}[t]
    \centering
    \includegraphics[trim = 0cm 0.25cm 0cm 0.25cm, clip=true, width=0.9\linewidth]{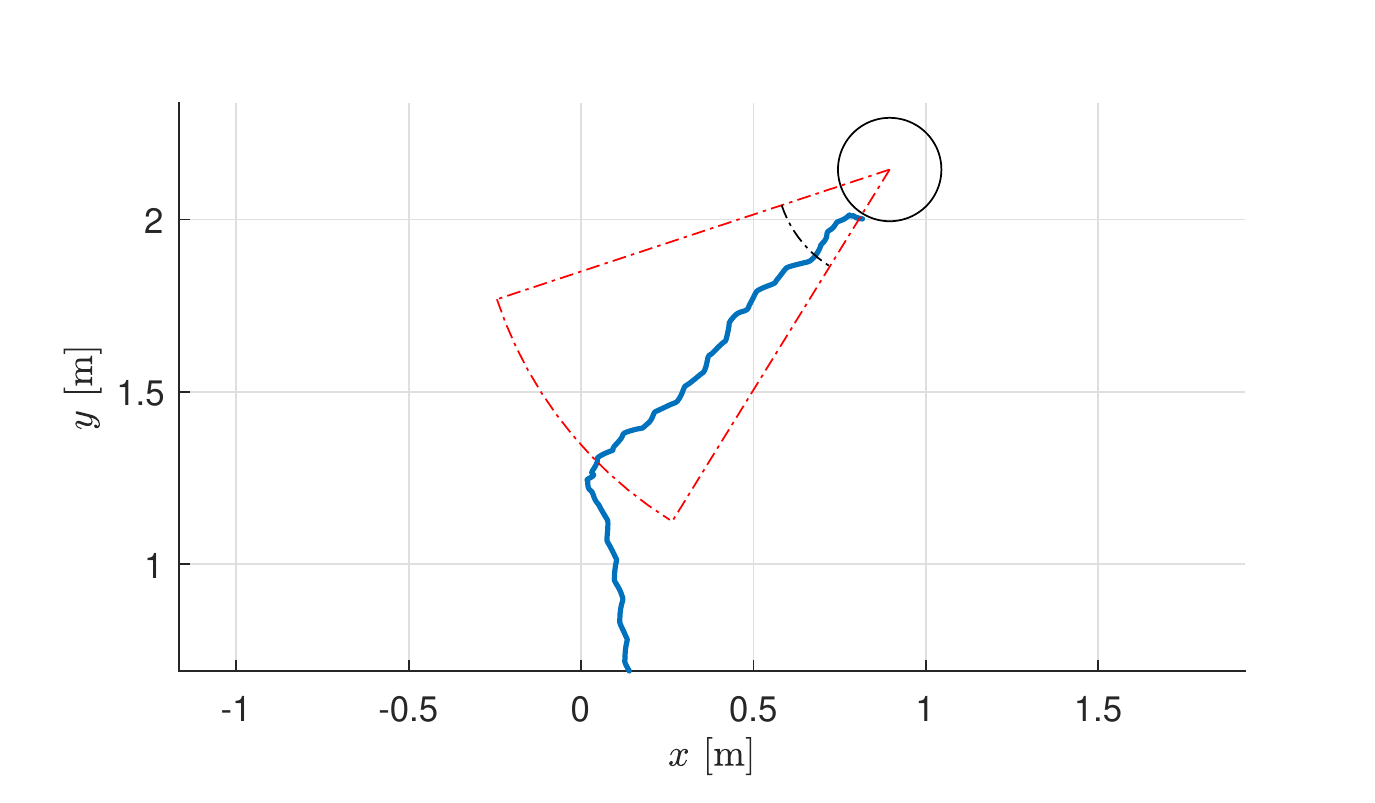}
    \caption{Sample trajectory (blue) for docking with a free-floating target. Dashed regions correspond to those in Fig.~\ref{fig:dockStrat}, but the target (black) is enlarged by the radius of the mobile boat. }
    \label{fig:trajSample}
\end{figure}
\begin{figure}[t]
    \centering
    \includegraphics[width=0.88\linewidth]{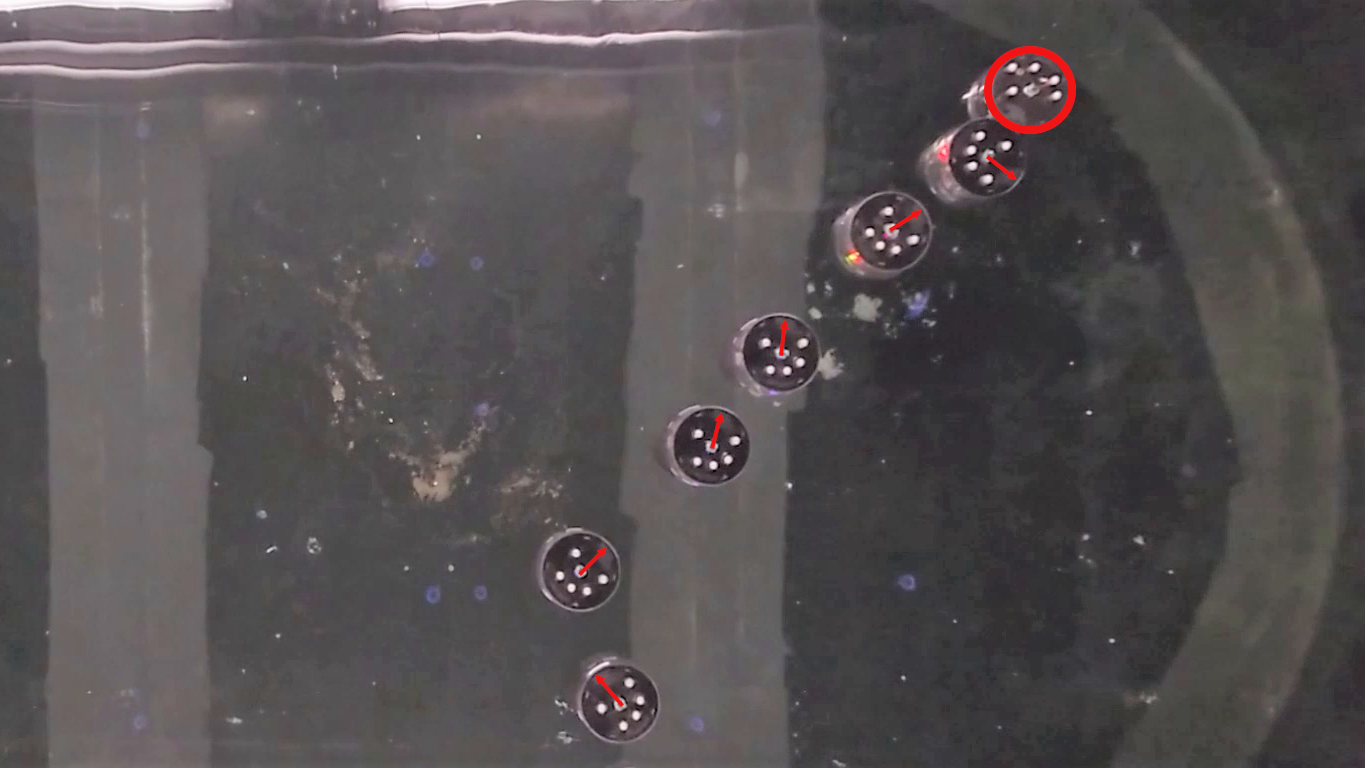}
    \caption{Composite of six snapshots corresponding to trajectory in Fig.~\ref{fig:trajSample}. The target boat and front of the mobile boat are highlighted in red. The orienting step occurs between the penultimate and last snapshots.}
    \label{fig:dockingComp}
\end{figure}

The results of the docking tests are presented in Table~\ref{tab:dockingExp}, and an example trajectory and composite image are shown in Figs.~\ref{fig:trajSample} and~Fig.~\ref{fig:dockingComp}, respectively. Front-docking and side-docking were performed reliably, while rear-docking was largely unsuccessful and so not attempted for the free-floating target. The time taken to dock from $0.5\si{m}$ is shown in Table~\ref{tab:dockingTimes}. Without loss of generality the desired target dock was set to $\psi_t = 0$ for all experiments, and the desired mobile dock $\psi_b$ was varied. The orientation transition velocity was set to $\omega_{tran} = 0.2\si{rad/s}$ and the approach distance to $d_{app} = 1.2\si{m}$. The approach cone was set to $\phi_{max} = 40^\circ$ wide.

Two measures of success are presented in Table~\ref{tab:dockingExp}. The first records whether the mobile boat hit the desired dock on the target $\psi_t$ with \textit{any} dock $\psi_b$, which measures the success of the approach method. The second records if the correct dock $\psi_b$ was used, which evaluates the orientation control method. If the mobile boat hits the wrong target dock $\psi_t$ the $\psi_b$ criterion is marked as a failure, regardless of which mobile dock was involved. 

The mobile boat was started at a random location in the experimental tank and allowed to run for 3 minutes or until a dock occurred, whichever came first. The desired angle $\theta_{des}$ maintained by orientation control was dynamically determined by the orientation of the line from the mobile boat to the target dock point. This allowed it to compensate for misalignment in the approach vector. 

\begin{table}[t]
    \centering
    \caption{Results of docking experiments}
    \begin{tabular}{c|c|c|c|c} \toprule
    & & \multicolumn{2}{c|}{Succ./Tries} & \\
    Target & $\hphantom{-}\psi_{b,des}$ & $\psi_t$ & $\psi_b$ & $d_{dock}$ [m] \\ \midrule
    \multirow{4}{*}{Static} & $\hphantom{-}0$ & 8 / 8  & 8 / 8 & 0.30 \\
    & $\hphantom{-}\pi/2$ & 7 / 8 & 7 / 8 & 0.30 \\
    &$-\pi/2$ & 8 / 8 & 8 / 8 & 0.30 \\
    &$\hphantom{-}\pi$ & 6 / 8 & 3 / 8 & 0.27 \\ \midrule
    \multirow{3}{*}{Free-Floating} & $\hphantom{-}0$ & 4 / 4 & 4 / 4 & 0.30 \\
    & $\hphantom{-}\pi/2$ & 3 / 3 & 3 / 3 & 0.33 \\
    & $-\pi/2$ & 2 / 2 & 2 / 2 & 0.33 \\\bottomrule 
    \end{tabular}
    \label{tab:dockingExp}
\end{table}

\begin{table}[t]
    \centering
    \caption{Docking times from $0.5\si{m}$ away, presented as $\mu \pm \sigma$}
    \begin{tabular}{c|c|c|c|c} \toprule
     & Approach [s] & Drift [s] & Latch [s] & Total [s]  \\ \midrule
    Modboats & $5.4 \pm 0.8$ & $2.5 \pm 1$ & $0$ & $8.0 \pm 1$ \\
    TEMP~\cite{OHara2014} & $8$ & $0$ & $4$ & $12$\\ \bottomrule 
    \end{tabular}
    \label{tab:dockingTimes}
\end{table}


\section{Discussion} \label{sec:discussion}

In our prior work~\cite{Knizhnik2020PausesRobot}, the Modboat required waypoints $20\si{cm}$ in radius for reasonable performance. In this work, however, the target waypoint (another boat) was only $7.6\si{cm}$ in radius. Nevertheless, our overall docking strategy and $40^\circ$ wide conical approach region allowed the Modboat to consistently approach and hit the target.  This can be seen in Table~\ref{tab:dockingExp}, where the success rate for hitting the target accurately (as indicated by docking with the correct $\psi_t$) is $97\%$ when rear-docks are excluded (and $84\%$ otherwise). 

This performance is due to a combination of factors aimed to combat the poor precision in the Modboat's low-cost and non-holonomic design. Discrete steering control provides poor position tracking, because the heading is calculated based on the current position but achieved at a different one; a wide conical approach provides plenty of room for the heading to converge. An observed downside of this approach is that heading towards the target does not drive the mobile boat away from the edges of the cone, which leads to occasional aborts due to drifting out of it. 

Additionally, the magnetic dock points are able to compensate for misalignment of the approach vector, and the dynamic choice of $\theta_{des}$ based on the mobile boat's location places the desired mobile dock in the most favorable orientation for capture. This allows the mobile boat to achieve initial contact at an offset point on the target, yet still be captured by the correct target dock $\psi_t$. This produces a $97\%$ success rate for front and side docking, which our method performs reliably.

In evaluating docking performance with a free-floating target, our chief concern was that the target might be pushed away by the oncoming mobile boat. Since we do not yet have a strategy for station-keeping with Modboats, this would mean that an alternative strategy would be needed to seed structures from single boats. In reality however, this effect is negligible; magnetic attraction is more prominent, and the free-floating target is able to rotate and translate to compensate for misalignment when a front-dock is attempted.

Free-floating targets do present an issue for side-docking, however. The orientation-control phase of the docking maneuver brings the mobile side-docks $\psi_b = \pm \pi/2$ close to the target, and if the approach vector is misaligned they may end up closer to the target side-docks than to the front-dock. For a fixed target this is not a problem, as orientation control will rotate through that configuration before magnetic attraction dominates. But if the target is able to rotate --- as the free-floating target is --- then it will do so and a side-to-side dock results ($\psi_t = \pm \pi/2$ and $\psi_b = \mp \pi/2$). We compensate by increasing the dock transition distance $d_{dock}$, as shown in the second half of Table~\ref{tab:dockingExp}. The extra space allows the orientation controller to rise to the desired orientation before magnetic forces are significant, and the dock occurs as before.

Rear-docking, however, has proven largely unsuccessful using our approach (Table~\ref{tab:dockingExp}). This occurs because the orientation-controlled maneuver to perform a $180^\circ$ rotation completely arrests the drift towards the target, and the oscillations that occur serve to then push the mobile boat away. This results in a lower $\psi_t$ success rate and a $\psi_b$ success rate below $50\%$. Lowering $d_{dock}$ can increase the chance of capture but leaves insufficient time to complete the maneuver, resulting in erroneous side-docks.

We can, however, still accomplish docking to all four dock points. As mentioned in Section~\ref{sec:undocking}, the tail can be used to rotate the dock point on the mobile boat by $\pm \pi/2$ when the target is docked to at least one other boat. This technique succeeded in 15 out of 16 attempts, so it can be considered a successful primitive for reorienting the mobile dock. Thus rear-docking can --- in all single-connection cases except seeding a structure --- be replaced by a front or side-dock followed by a tail-based reorientation. It must be used carefully, however, since we have little control during the transition and can expect to encounter issues when external flows are present.

Docking times following our strategy compare favorably to those displayed by the TEMP project~\cite{OHara2014} as shown in Table~\ref{tab:dockingTimes}. Although we introduce an additional drift phase for orientation control, time is saved due to the lack of active latching. This allows the Modboat to dock in a comparable time frame to other modular systems.

\section{Conclusions} \label{sec:conclusion}

In this work we have presented a design and a strategy for docking the Modboat --- a single-motor, oscillating, surface-swimming robot --- to other Modboats. We have also demonstrated that the robot is capable of undocking from a docked position by using its tail to break the magnetic connections to its neighbors. Together, these results allow Modboats to self-assemble and reconfigure into potentially large structures following a 4-connected lattice. This docking and undocking is accomplished with only one  actuator per module --- without the need for additional actuators for (un)docking --- which the authors believe is the first instance for any modular self-reconfigurable robot system.

We have also developed an additional motion-primitive for the Modboat, which allows control of the instantaneous orientation of the top body. In combination with our docking strategy, this enables the Modboat to dock in either a front-facing or a side-facing configuration despite being able to control its orientation only in the average while swimming. This method cannot reliably achieve successful rear-facing docks, but we can achieve rear-docks by side-docking and then reorienting using the tail. The combination of these methods allows Modboats to achieve a full set of potential configurations, with applications for orientation-sensitive payloads such as flow sensors.

In this work we have considered docking to/undocking from only a single target. Docking to multiple modules simultaneously will present further challenges, as multiple magnetic links must be made simultaneously; undocking will similarly be more difficult when multiple connections exist, and may not be possible for all configurations. Other modules in the structure may also exert magnetic disturbance forces during the approach, which will need to be accounted for before large structures can be assembled. We plan to explore these challenges in future work.

We also plan to explore steering functionality to drive the Modboat towards the center of the approach cone and decrease the number of retries needed, and to consider the effects of disturbances such as from external flows. We have also verified experimentally that the docking mechanism described in this work is sufficient to allow multiple Modboats to swim together, but this type of motion has not been explored. In future work we will consider the various types of motion allowed by two or more Modboats swimming while linked together in various orientations.


\section*{Acknowledgment}

We thank Dr. M. Ani Hsieh for the use of her instrumented water basin in obtaining all of the testing data.

\bibliographystyle{./bibliography/IEEEtran}
\bibliography{./bibliography/IEEEabrv,./bibliography/iros2020,./bibliography/nonpaper,./bibliography/references}

\end{document}